\documentclass[conference]{IEEEtran}
\IEEEoverridecommandlockouts
\usepackage{cite}
\usepackage{amsmath,amssymb,amsfonts}
\usepackage{algorithmic}
\usepackage{graphicx}
\usepackage{textcomp}
\usepackage{xcolor}
\usepackage{svg}
\usepackage{orcidlink}
\usepackage{booktabs} 
\usepackage{xcolor,colortbl} 
\usepackage{subcaption}
\usepackage{float}
\usepackage[margin=1in]{geometry}
\usepackage{stfloats}
\usepackage{xcolor}
\usepackage{longtable}
\usepackage{array}
\usepackage{multirow}
\usepackage{array,booktabs}
\def\BibTeX{{\rm B\kern-.05em{\sc i\kern-.025em b}\kern-.08em
    T\kern-.1667em\lower.7ex\hbox{E}\kern-.125emX}}

\usepackage{fancyhdr}
\usepackage{geometry}
\usepackage{lipsum} 

\begin{document}

\title{Enhancing Safety Standards in Automated Systems Using Dynamic Bayesian Networks}

\author{
\IEEEauthorblockN{
Kranthi Kumar Talluri\IEEEauthorrefmark{1}, Anders L. Madsen\IEEEauthorrefmark{2}\IEEEauthorrefmark{4}, Galia Weidl\IEEEauthorrefmark{1} 
}

\IEEEauthorblockA{\IEEEauthorrefmark{1}Aschaffenburg University of Applied Sciences, Aschaffenburg, 63743 Germany}
\IEEEauthorblockA{\IEEEauthorrefmark{2}HUGIN EXPERT A/S}
\IEEEauthorblockA{\IEEEauthorrefmark{4}Department of Computer Science, Aalborg University, Denmark}
\IEEEauthorblockA{Email: 
\IEEEauthorrefmark{1}\{KranthiKumar.Talluri, Galia.Weidl\}@th-ab.de,
\IEEEauthorrefmark{2}anders@hugin.com}

\thanks{We are grateful to the Smart Mobility Project at the European Commission, Joint Research Center and in particular to Konstantinos Mattas and Biagio Ciuffo for introducing us to the regulations, to the related open research questions, and for sharing with us the repository of the code, which we used to integrate our model for comparison \cite{jrc-fsm}.}

}

\maketitle

\begin{abstract}

Cut-in maneuvers in high-speed traffic pose critical challenges that can lead to abrupt braking and collisions, necessitating safe and efficient lane change strategies. We propose a Dynamic Bayesian Network (DBN) framework to integrate lateral evidence with safety assessment models, thereby predicting lane changes and ensuring safe cut-in maneuvers effectively. Our proposed framework comprises three key probabilistic hypotheses (lateral evidence, lateral safety, and longitudinal safety) that facilitate the decision-making process through dynamic data processing and assessments of vehicle positions, lateral velocities, relative distance, and Time-to-Collision (TTC) computations. The DBN model's performance compared with other conventional approaches demonstrates superior performance in crash reduction, especially in critical high-speed scenarios, while maintaining a competitive performance in low-speed scenarios. This paves the way for robust, scalable, and efficient safety validation in automated driving systems. 

\end{abstract}

\begin{IEEEkeywords}
Dynamic Bayesian Network, Probabilistic Risk Assessment, Autonomous Driving Safety, Lane Change Prediction.
\end{IEEEkeywords}

\section{Introduction}

The presence of advanced autonomous vehicles(AVs) in real-world traffic is increasing daily, necessitating the need for robust models that can estimate risks and plan maneuvers proactively to ensure safety. Accurate detection and prediction of lane change maneuvers are crucial for collision avoidance, traffic flow optimization, and safety enhancement \cite{lu2024studying}. Cut-ins create dynamic and unpredictable traffic conditions, leading to unforeseen braking and lower safety margins.  

Traditional heuristic rule-based models and statistical frameworks are employed in numerous research studies to understand lane changes. However, they fail to cope with dynamic traffic environments and real-world uncertainties due to their dependency on predefined conditions. AI/ML advancements facilitate the use of techniques like deep learning (DL) and reinforcement learning (RL)-based approaches for lane change prediction. Methods like Deep reinforcement learning (DRL) integrated with a finite state machine (FSM) \cite{shen2024comfortable}, Model Predictive Control (MPC), and Inverse RL have been applied for optimal Adaptive Cruise Control (ACC) through lateral movement awareness, trajectory optimization, and driving style adaption, respectively \cite{paul2024autonomous, qiu2024driving}. Some research works have used transformer-based models to integrate driver gaze behavior for recognizing lane changes \cite{li2024lane}. These advancements laid an effective groundwork, but several limitations still persist. Deep learning models heavily rely on perfect perception and are challenging to apply in real-world applications due to environmental uncertainties \cite{zhang2022long}. The RL-based framework handled emergency maneuvers by maintaining vehicle stability at the cost of high computational demands \cite{shen2024comfortable, zhao2024autonomous}, and explainability due to its black box nature. Furthermore, cooperative lane change strategies and MPC-based statistical models compromise the traffic throughput and fail to adapt to unforeseen human behaviors. These limitations restrict the model's usage for real-world applications \cite{dong2024adaptive}.

Probabilistic models such as Bayesian Networks (BNs) help overcome some of these challenges \cite{weidl2018early}. BNs are powerful tools for real-time decision-making because they are interpretable and can model uncertainties. Furthermore, BNs excel at integrating multi-sensor data along with human domain knowledge to predict situations more accurately. 

Our earlier work demonstrated that Object-Oriented Bayesian networks (OOBNs) effectively predict maneuver recognition in real highway scenarios \cite{Kasper2011, kasper2012object, kasper2012erkennung}. We integrated parameters related to vehicle kinematics, interaction with surrounding vehicles, and estimates of driver intent, among others, and obtained promising results \cite{Kasper2011, kasper2012object, kasper2012erkennung}. We further extended our OOBNs model to DBNs and deployed it on an experimental car, successfully testing it in real-world critical highway traffic scenarios for lane change detection \cite{weidl2014optimizing, weidl2015early, weidl2018early, weidl2016situation}. Taking advantage of our previous findings, we enhance DBN's safety assessment in our current work for effective lane change detection and cut-in maneuver planning in the JRC-FSM simulator \cite{mattas2020fuzzy, jrc-fsm}(simulator is detailed in section \ref{sec:Simulator}). The performance of the proposed DBN is enhanced through the integration of revised safety components, which not only help handle cut-in scenarios better but are also robust against high-speed lane changes and rough vehicle interaction situations. The contributions of this work are listed: 

\begin{itemize} 
    \item Extension of the existing DBN framework for effective risk assessment through updated probabilistic safety checks.
    \item The DBN is modeled by integrating the lateral evidence of lane change into the JRC-FSM simulator's baseline model named CC\_human\_driver for proactive safety checks. 
    \item The Proposed framework is evaluated using metrics such as TTC, jerk, and deceleration for various low- and high-speed scenarios.
    \item Comparison of DBN model's performance against CC\_human\_driver, FSM, RSS, and Reg157 models.
\end{itemize}

The rest of the paper is organized as follows: Section II provides background information about the DBN and the JRC-FSM simulator. Section III provides a brief description of maneuver recognition modeling. Section IV discusses the performance evaluation of the proposed framework, and the conclusion summarizes its advantages and future plans.


\section{Background Information}

\subsection{Fundamentals of BN}

A BN is a probabilistic model capable of effectively representing complex relationships between a set of variables and their conditional dependencies through a Directed Acyclic Graph (DAG) \cite{kjaerulff2008bayesian}. Variables are represented as nodes, and their dependencies as edges. Their joint probability distribution decomposes as:

\begin{equation}
P(X_1, X_2, \ldots, X_n) = \prod_{i=1}^n P(X_i \mid \text{Pa}(X_i)),
 \label{eq:bn_eq1}
\end{equation}

where $\text{Pa}(X_i)$ is the parent nodes of $X_i$ in Equation \ref{eq:bn_eq1}. 
The posterior probabilities are computed based on evidence obtained through Bayes theorem formulated as shown in Equation \ref{eq:Bayestheorem}, which is a fundamental concept of BN.

\begin{equation}
P(A \mid B) = \frac{P(B \mid A) \cdot P(A)}{P(B)}.
\label{eq:Bayestheorem}
\end{equation}

$A$ is the hypothesis, $B$ is the evidence and  $P(B \mid A)$ represents the likelihood of occurrence of $B$ when a hypothesis is known \cite{talluri2024bayesian, talluri2024impact}.



\subsection{Simulator Framework}
\label{sec:Simulator}
\begin{figure}[H]
    \centering
    \includegraphics[width=\linewidth]{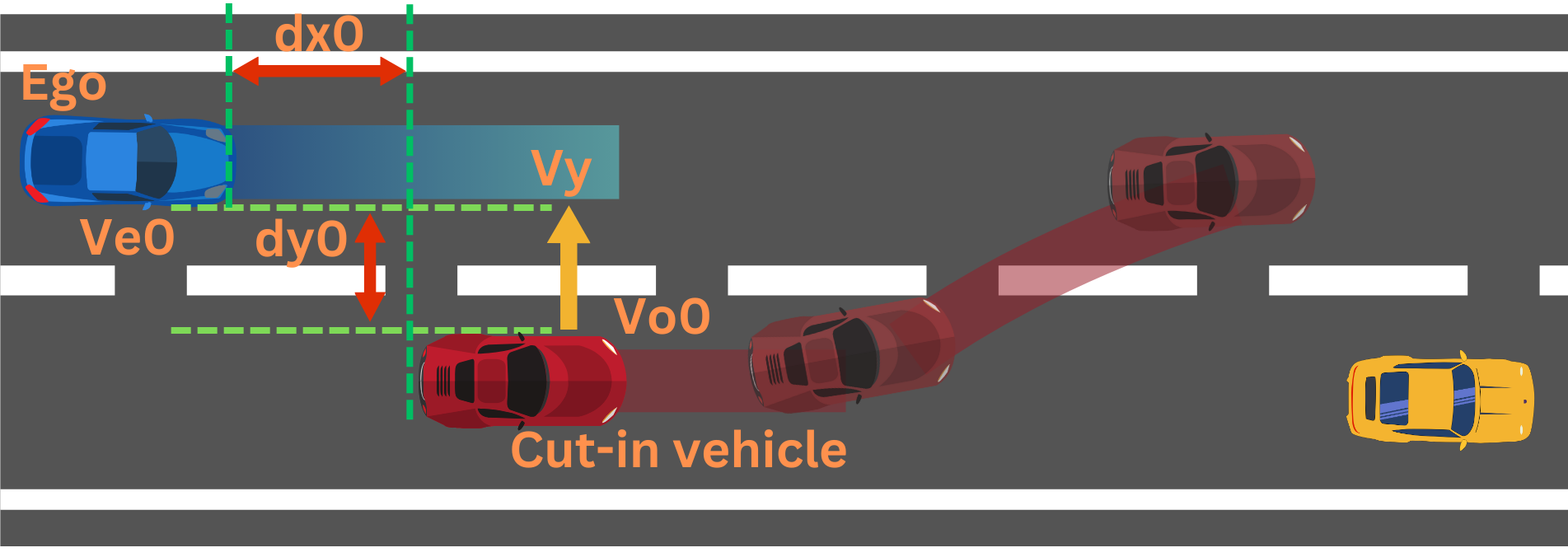} 
    \caption{Variable notations for ego and cut-in vehicle in JRC-FSM framework}
    \label{fig:JRC-FSM_framework} 
\end{figure}

The Joint Research Centre's Fuzzy Safety Model (JRC-FSM) \cite{mattas2020fuzzy} simulation framework was developed to strictly evaluate the effectiveness of developed safety cut-in algorithms under complex cut-in scenarios. Earlier developed models like the Fuzzy Safety Model (FSM), Responsibility-Sensitive Safety (RSS) \cite{shalev2017formal}, CC\_human\_driver \cite{UNECEReg157}, and Reg157 \cite{UNECEReg157} are integrated into this Python-based JRC-FSM framework. The key parameters of the Cut-in maneuver in the simulator are shown in Figure \ref{fig:JRC-FSM_framework} and Table \ref{tab:Parameters_description}.

\begin{table}[h!]
\centering
\caption{Parameters description of ego and cut-in}
\label{tab:Parameters_description}
\begin{tabular}{cc}
\hline
\textbf{Parameter} & \textbf{Description} \\ \hline
$dx0$ & Longitudinal distance between ego and cut-in \\ 
$dy0$ & Lateral distance between ego and cut-in \\ 
$Ve0$ & Velocity of ego \\ 
$Vy$ & Lateral velocity \\ 
$Ve0 - Vo0$ & Relative velocity between ego and cut-in vehicle \\ \hline
\end{tabular}
\end{table}

Simulations are tailored extensively and precisely to particular cut-in through various parameter combinations of initial ego speed, cut-in speed, lateral velocity, and initial distance. Accurate modeling of cut-in scenarios is crucial, as they have a significant impact on evaluation outcomes. Furthermore, the JRC-FSM framework supports systematic evaluations of individual test cases. The framework facilitates seamless integrations and comparisons of advanced safety models, making it a robust tool for expanding the safety modeling field.


\section{Modeling the recognition of Lane Change maneuver}


DBN  is used to model systems that evolve over time. The state of the system at a single time instant is modeled using an ordinary static Bayesian network. By linking the networks at each consecutive time slice, a dynamic model is obtained that handles the input-output relation. DBN infers the probability of safety-critical events using real-time inputs, such as positions, velocities, and TTC, as evidence, and produces safety decisions as output at each time step.

The DBN is built using the commercial HUGIN software, which allows efficient modeling and deployment into the JRC-FSM simulator. DBN and its integration into the simulator mainly focus on improving the CC\_human\_driver baseline model through understanding and analyzing its limitations in proactive detection and seamless response to the cut-in possibilities. The primary functionalities of each model are safety\_check and react. This work focuses on enhancing the safety\_check function using lateral evidence for collision avoidance strategy, which uses key parameters like lateral velocity, longitudinal distance, relative velocity, and TTC to effectively detect the situation without causing a crash. While modeling, the DBN is set to use standard safety parameters, as shown in Table \ref{tab:Safety Parameters}. 

\begin{table}[h!]
\centering
\caption{Safety Parameters of CC\_human\_driver}
\label{tab:Safety Parameters}
\begin{tabular}{ccc}
\hline
\textbf{Parameter} & \textbf{Description} & \textbf{Values}\\ \hline
$cc_{rt}$ & Reaction time & 0.75 \\ 
$cc_{min\_jerk}$ & Minimum jerk & 12.65 \\ 
$cc_{max\_deceleration}$ & Max deceleration & 0.774 * 9.81 \\ 
$cc_{release\_deceleration}$ & Deceleration release & 0.4 \\ 
$cc_{critical\_ttc}$ & Critical TTC & 2 \\ \hline
\end{tabular}
\end{table}

The proposed DBN consists of 3 key probability hypotheses, namely LE, safe\_lat, and safe, which ensures early lane change recognition and seamless cut-in maneuvers through the listed DBN's interconnected workflow, which is the complete cycle for every cut-in scenario. The mathematical formulas of CC\_human\_driver for the lateral and longitudinal safety assessment are modeled in safe\_lat, and safe \cite{UNECEReg157}.

\begin{figure}[h]
    \centering
    \includegraphics[width=1\linewidth, height=0.17\textheight]{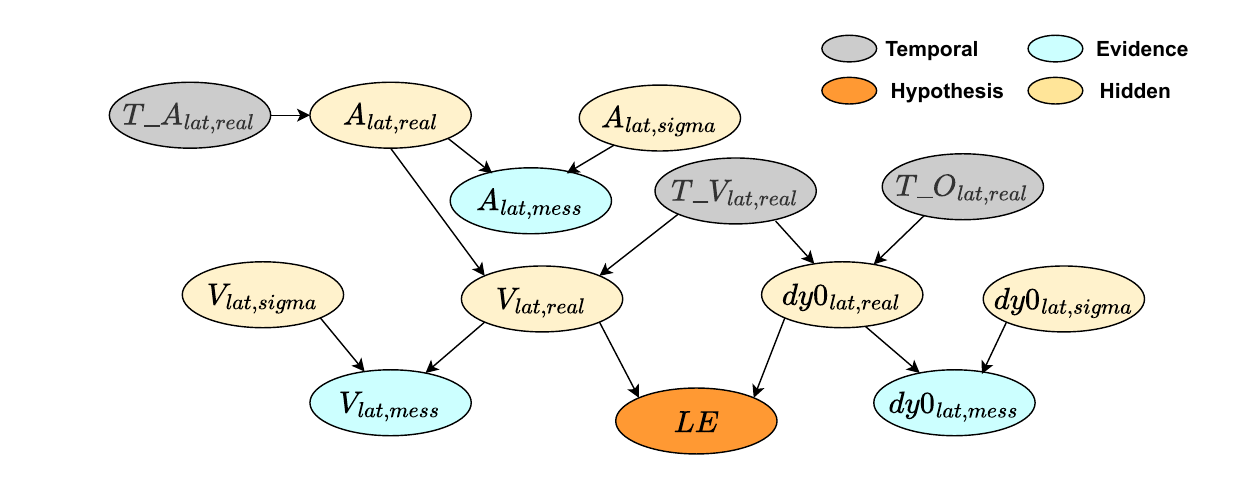}
    \caption{LE providing the lateral evidence for lane change recognition.}
    \label{fig:LE_block} 
\end{figure}

Initially, the input data with information like lateral velocity ($V_{lat, real}$) and lateral distance ($dy0_{\text{lat}}$) among the ego and cut-in vehicle is analyzed by \textbf{LE} as shown in Figure \ref{fig:LE_block} and all corresponding parameters are listed in Table \ref{tab:parameters_ego_cut-in}. The sigmoid function shown in the equation represents the CPD of LE, which expresses the increasing probability of LE (indicating the possibility of lane change) when a cut-in vehicle approaches closer to the ego vehicle modeled through node $dy0_{lat,mess}$ and by increasing lateral velocity modeled in node $V_{lat,mess}$. This sigmoid curve is optimized using Python libraries with sample data collected from the cut-in scenario, where the ego vehicle speed is $100 \, \text{km/h}$ and the cut-in vehicle speed is $60 \, \text{km/h}$. The sigmoid function is mathematically formulated as: 


\begin{align}
P(LE = \text{true} \mid V_{\text{lat}}, dy0_{\text{lat}}) &= 
\frac{s_v}{s_v + e^{m_v \cdot V_{\text{lat}}}} \cdot \nonumber \\
& \quad \frac{s_o}{s_o + e^{m_o \cdot dY0_{\text{lat}}}},
\end{align}

The values $s_v = 0.062$, $m_v = 8.945$, $s_o = 3.386$, and $m_o = 7.313$ were obtained from optimized sigmoid curve. For dynamic inference during lane change, smooth sigmoid-based probabilistic mapping is crucial to guarantee accurate predictions. The obtained equation is used in LE to compute lane change probabilities.

\begin{align}
P_1 &= \frac{0.062}{0.062 + \exp(8.945 \cdot V_{lat, real})}, \\
P_2 &= \frac{3.386}{3.386 + \exp(7.313 \cdot dy0_{lat, real})}
\end{align}


\begin{align}
P(LE) = 
\begin{cases}
1 - \min(P_1 \cdot P_2, 1), &  \text{false} \\
\min(P_1 \cdot P_2, 1), & \text{true}
\end{cases}
\end{align}

\begin{table}[h!]
\centering
\caption{Parameters of ego and Cut-in}
\label{tab:parameters_ego_cut-in}
\begin{tabular}{cc}
\hline
\textbf{Parameter} & \textbf{Description} \\ \hline
$pos_{ego, lat}$ & Lateral position of ego \\ 
$pos_{cut-in, lat}$ & Lateral position of cut-in \\ 
$l_{ego}, w_{ego}$ & Length and width of ego \\ 
$l_{cut-in}, w_{cut-in}$ & Length and width of cut-in \\ 
$pos_{ego, long}$ & Longitudinal position of ego\\ 
$pos_{cut-in, long}$ & Longitudinal position of cut-in \\ 

$V_{ego, long}$ & Longitudinal speed of ego \\ 
$V_{cut-in, long}$ & Longitudinal speed of cut-in \\ 
 \hline
\end{tabular}
\end{table}

\begin{figure}[H]
    \centering
    \includegraphics[width=0.8\linewidth, height=0.11\textheight]{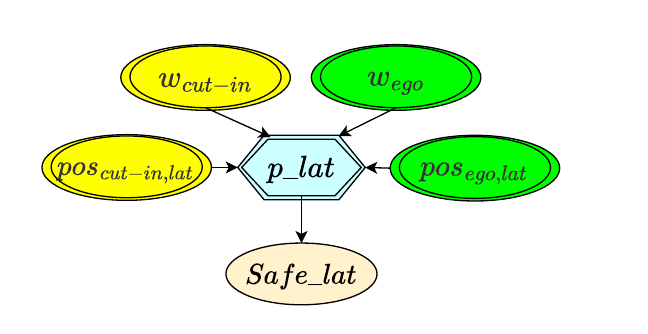} 
    \caption{Architecture of lateral Safety.}
    \label{fig:Safe_lat} 
\end{figure}

The safety assessments are further refined by \textbf{Safe\_LAT} as shown in Figure \ref{fig:Safe_lat}. It is responsible for evaluating the lateral safety margin depending on the position and width of the ego and cut-in vehicle. It confirms whether lateral maneuvers are under safe thresholds by calculating and analyzing if there is sufficient lateral clearance (\(p_{\text{lat}}\)) between both vehicles \cite{UNECEReg157}. 

\begin{align}
p_{lat} = 
\begin{cases}
1, & \text{if } (pos_{ego, lat} - pos_{cut-in, lat} \\
& \quad - \frac{w_{ego}}{2} - \frac{w_{cut-in}}{2}) > 0, \\
0, & \text{otherwise}
\end{cases}
\end{align}


\begin{align}
Safe_{lat} = 
\begin{cases}
1-p_{lat}, &  \text{false} \\
p_{lat}, & \text{true}
\end{cases}
\end{align}

\begin{figure}[H]
    \centering
    \includegraphics[width=1\linewidth, height=0.15\textheight]{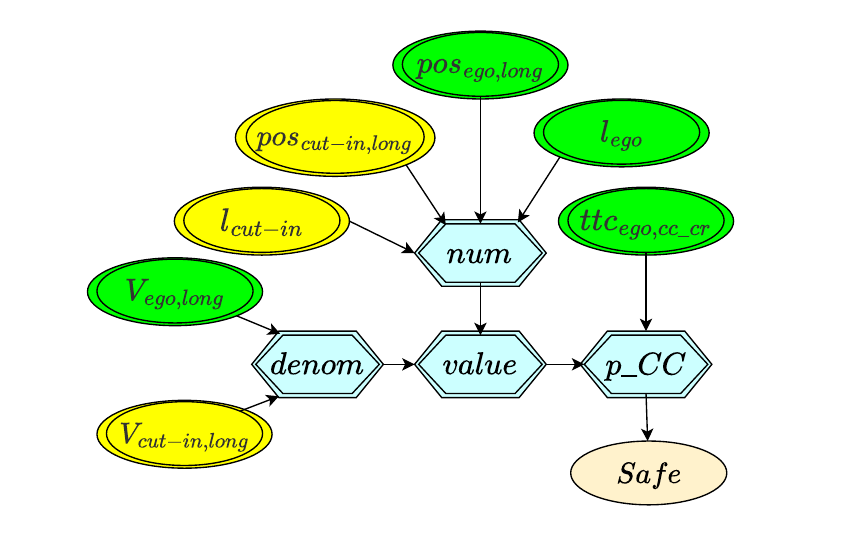} 
    \caption{Architecture of longitudinal Safety.}
    \label{fig:safe} 
\end{figure}

The longitudinal safety is evaluated in parallel by \textbf{Safe} hypotheses as shown in Figure \ref{fig:safe}, which measures significant parameters like minimum Time-To-Collision (TTC), relative speed, and longitudinal distance (\(dx_0\)). They use the connection between nodes describing the length ($l$) and position ($pos$) of ego and cut-in vehicle to compute \(p_{\text{CC}}\) and evaluate the likelihood of a collision occurrence \cite{UNECEReg157}. The DBN model utilizes in the HUGIN tool some function nodes, in which features are specified by the following expressions: 

\begin{align}
\text{num} &= |pos_{cut-in, long} - pos_{ego, long} \nonumber \\
& \quad - \frac{l_{ego} + l_{cut-in}}{2}|, \\
\text{denom} &= pos_{ego, long} - pos_{cut-in, long}, \\
\text{value} &= 
\begin{cases}
\frac{\text{num}}{\text{denom}}, & \text{if } \text{denom} \neq 0, \\
0, & \text{otherwise}
\end{cases}
\end{align}

If an unsafe situation is predicted, the Safe hypothesis promptly calculates the required deceleration and speed adjustments. The maximum deceleration and jerk limits are set to \(0.774 * 9.81 \, \text{m/s}^2\) and \(12.65 \, \text{m/s}^3\), respectively, to ensure that the cut-in scenarios and possible responses are realistically achievable. As the JRC-FSM simulator is designed to operate at a 10 Hz update rate, enabling real-time safety evaluation, for continuous monitoring and real-time adaptation, the DBN is also set to operate at a frequency of 10 Hz, i.e., every 100 ms.

\begin{align}
p_{CC} &= 
\begin{cases}
1, & \text{if } \text{value} > ttc_{ego, cc\_cr}, \\
0, & \text{otherwise}.
\end{cases}
\end{align}

\begin{align}
Safe = 
\begin{cases}
1-p_{CC}, &  \text{false} \\
p_{CC}, & \text{true}
\end{cases}
\end{align}

The dynamic nodes represented as ovals in grey color are updated in real-time depending on incoming data. The probabilistic distributions of these nodes are dynamically adjusted at \(t = 15\) (15 time slices for temporal information) using the vehicle dynamics, specifically positions, velocities, and accelerations, which exhibit the evolving cut-in scenario. Moreover, analysis of these vehicle dynamics helps DBN in future trajectory prediction by computing the time to boundary (ttb) and future positions to ensure safety with the following equations:

\begin{equation}
\text{ttb} = \frac{dy_0}{-V_y}
\end{equation}

\begin{equation}
\begin{aligned}
\Delta x_{\text{obj}} &= V_{o0} \cdot \text{ttb}, \quad \Delta x_{\text{ego}} = V_{e0} \cdot \text{ttb}, \\
x_{\text{pred, ego}} &= \Delta x_{\text{ego}} + x_{\text{ego}}, \quad x_{\text{pred, obj}} = \Delta x_{\text{obj}} + x_{\text{obj}}, \\
x_{\text{pred, ego}} &> x_{\text{pred, obj}}
\end{aligned}
\end{equation}

Dynamic updates and evaluation of vehicle states, trajectory prediction, and safety parameters provide effective risk evaluation, reduced crash risks, and smooth vehicle control to mitigate collision possibilities.

\begin{table}[h!]
\centering
\caption{LE Nodes and State Discretization}
\label{tab:LE_Nodes_and_states}
\centering
\begin{tabular}{ll}
\hline
\textbf{Node} & \textbf{State Discretization} \\
\hline
$dy0_{lat, real}$ & $-2.0,..., 2.0$ m (0.1 m steps) \\
$dy0_{lat, mess}$ & $-1,..., 1.9$ (0.1 m steps)\\
$V_{lat, real}$ & $-1.9,..., 0.5$ m/s (0.1 m/s steps) \\
$V_{lat, mess}$ & $-1.8,..., 0.4$ (0.1 m/s steps) \\
$dy0_{lat, sigma}$ & $0 - 0.001, 0.001 - 0.05, 0.05 - 0.1, 0.1 - 0.2$ \\
$V_{lat, sigma}$ & $0 - 0.001, 0.001 - 0.05, 0.05 - 0.1, 0.1 - 0.2$ \\
$A_{lat, sigma}$ & $0,..., 0.125$ (0.025 steps) \\
$A_{lat, real}$ & $0,..., 1.5$ m/s$^2$ (0.1 m/s$^2$ steps) \\
$A_{lat, mess}$ & $-0.001,..., 1.5$ (0.1 steps) \\
\hline
\end{tabular}
\end{table}

The DBN model operates with both continuous and discrete variables. While Safe and Safe\_LAT hypotheses utilize continuous variables directly from the JRC-FSM simulator \cite{jrc-fsm}, the LE hypotheses contain discretized state variables with intervals as shown in Table \ref{tab:LE_Nodes_and_states}. The DBN is implemented using various modeling techniques, including the fundamental rule, the chain rule, and Bayes' theorem with conditional probability tables (CPTs) for discrete variables and dynamic parameter updates.


\section{Results and Discussion}

This section compares the performance of the proposed DBN model with other models for various low- and high-speed scenarios. 

\subsection{Enhancing the CC\_human\_driver Model}

\begin{table*}[h!]
\renewcommand{\arraystretch}{1.2}
\centering
\caption{Comparison of Crash Avoidance Rate and Average Minimum TTC at different Ego Speeds.}
\label{tab:Crash_Avoidance_and_TTC}
\begin{tabular}{c|ccc|ccc}
\hline
\textbf{Ego Speed (km/h)} & \multicolumn{3}{c|}{\textbf{Crash Avoidence Rate}} & \multicolumn{3}{c}{\textbf{minTTC}} \\ 
                           & \textbf{CC\_human\_driver (\%)} & \textbf{DBN (\%)} & \textbf{Improved (\%)} & \textbf{CC\_human\_driver (s)} & \textbf{DBN (s)} & \textbf{Improved (s)} \\ \hline
\centering 70  & 80.0 & 95.4 & +15.4 & 3.00 & 1.00 & +2.00 \\
\centering 90  & 79.5 & 94.5 & +15.0 & 2.80 & 0.87 & +1.93 \\
\centering 110 & 78.0 & 96.7 & +18.7 & 2.00 & 0.96 & +1.04 \\
\centering 130 & 77.0 & 96.0 & +19.0 & 1.10 & 0.58 & +0.52 \\ \hline
\end{tabular}
\end{table*}

Lane change detection and crash avoidance through integrating lateral evidence (LE) into the CC\_human\_driver model is a necessary advancement for effective improvement in safety and reliability. The LE hypothesis enhances lane change detection by predicting the probability of a cut-in vehicle's intention to change lanes at least 1.5 seconds earlier than the CC\_human\_driver model, which reduces the likelihood of a crash. At all high-speed scenarios, the DBN model shows superior performance in crash avoidance with 15-19\% more than CC\_human\_driver across all ego speeds as shown in Table \ref{tab:Crash_Avoidance_and_TTC}. Furthermore, Table \ref{tab:Crash_Avoidance_and_TTC} also highlights the DBN's capability to improve the average minimum TTC. When compared to the CC\_human\_driver model, DBN adds an improvement of +2.00 seconds TTC at low speeds and +0.52 seconds TTC at high speeds. DBN aims to prevent unforeseen or unsafe vehicle behaviors while maintaining a safety margin, even in critical high-speed traffic conditions. 

\begin{figure}[h]
    \centering
    \includegraphics[width=0.8\linewidth, height=0.15\textheight]{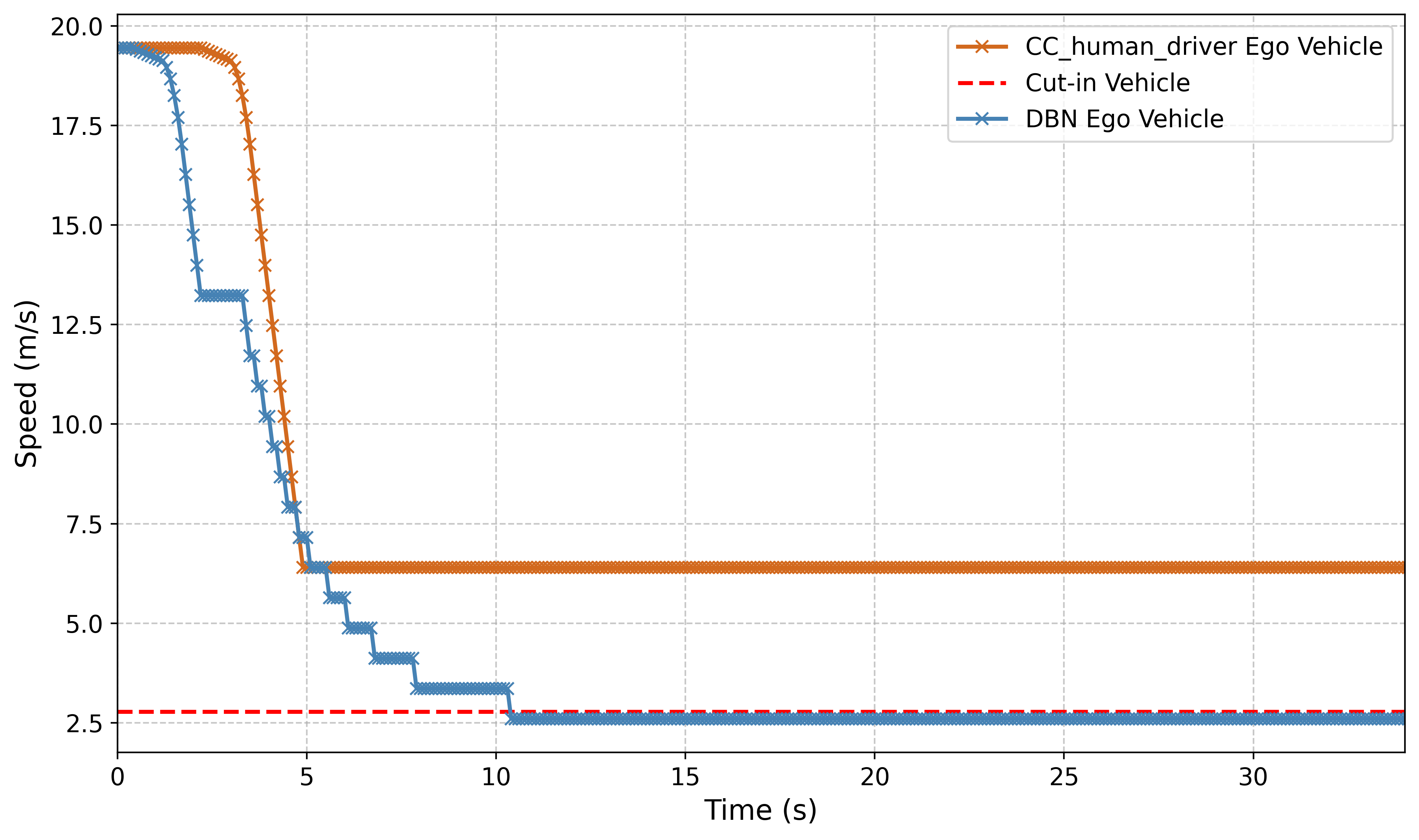} 
    \caption{Braking capabilities of DBN and CC\_human\_driver for a particular high speed scenario.}
    \label{fig:Speed_profiles} 
\end{figure}

For a better understanding of the DBN's proactive and efficient recognition of lane changes, a scenario specifically designed for high-speed environments is selected, demonstrating the braking capabilities of the DBN and CC\_human\_driver models, as shown in Figure \ref{fig:Speed_profiles}. In this scenario, we set the ego vehicle and cut-in vehicle speeds at 70 km/h and 10 km/h, respectively. It can be seen that the DBN detects the lane change approximately 3 seconds earlier than the CC\_human\_driver model, which helps the DBN brake smoothly and cause less jerk, a crucial factor for comfortable driving. This braking is applied until the speed reaches below the cut-in speed (the dotted red line) and follows the cut-in vehicle without causing a crash. On the contrary, with the CC\_human\_driver model, the ego vehicle crashes into the cut-in vehicle as it fails to brake in time. 

\begin{figure}[h]
    \centering
    \includegraphics[width=0.9\linewidth, height=0.20\textheight]{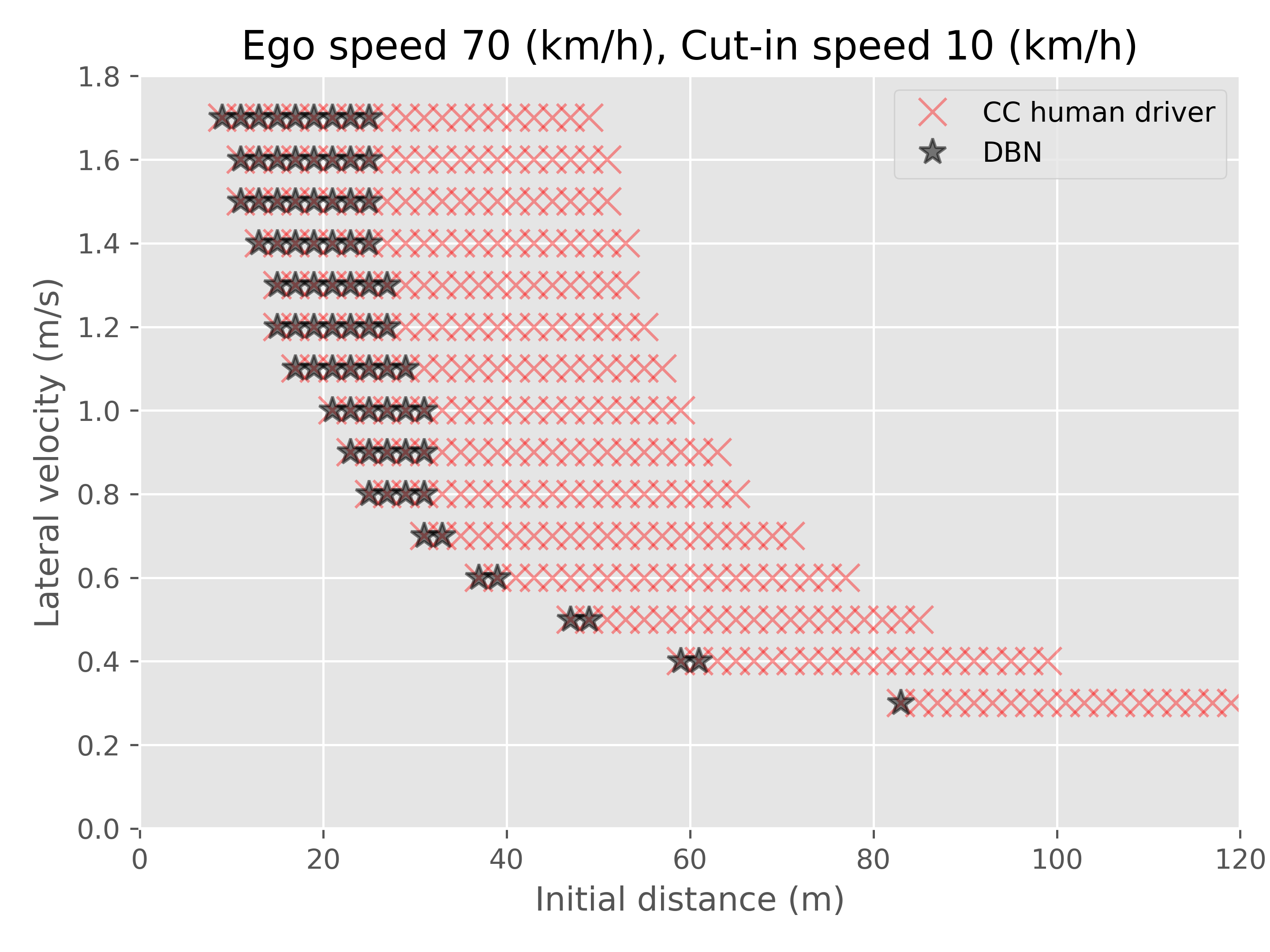} 
    \caption{Comparing the improved performance of DBN with CC\_human\_driver model}
    \label{fig:improved_performance_DBN} 
\end{figure}

One significant aspect of enhancing a particular model is to minimize the existing crashes while preventing the occurrence of additional crashes. To illustrate this aspect, we select one of the critical scenarios with a high relative speed difference, where the ego vehicle speed is 70 km/h and the cut-in speed is 10 km/h across all combinations of lateral distance and initial distance. As seen in Figure \ref{fig:improved_performance_DBN}, DBN effectively avoids more crashes and focuses solely on reducing the crashes of the CC\_human\_driver model. These results highlight the advantage of using probabilistic models to achieve a safer and more predictable automated driving experience, thereby strengthening the trustworthiness of using DBNs for tackling real-world critical scenarios.  

\subsection{Comparative Performance of the DBN Model}

\begin{figure*}[h]
    \centering
    \begin{subfigure}[b]{0.328\textwidth}
        \centering
        \includegraphics[width=\textwidth]{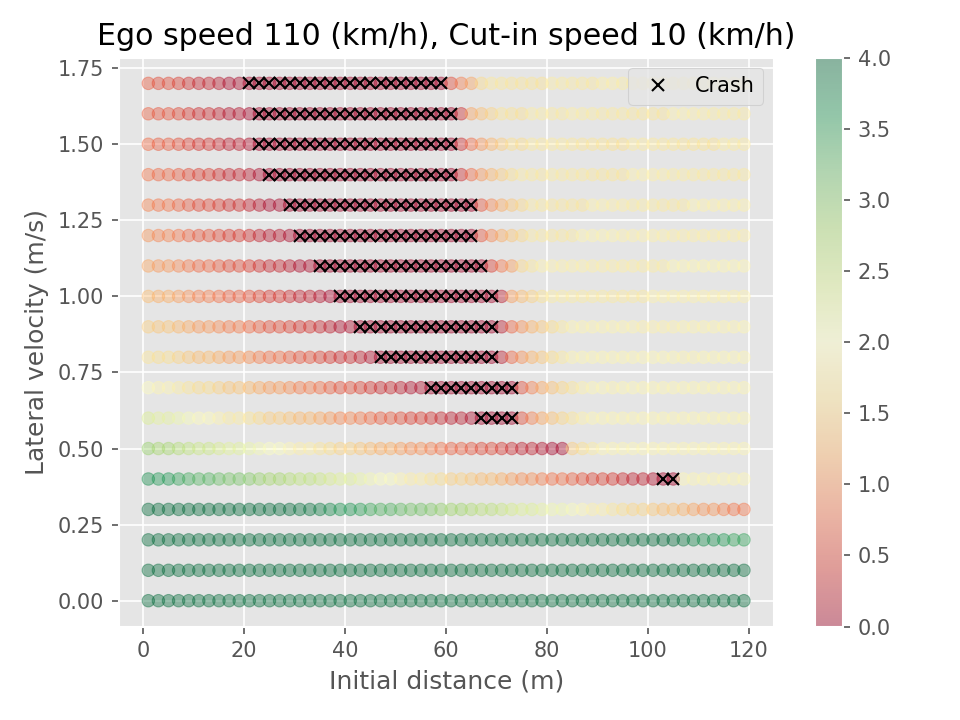}
        \caption{DBN}
        \label{fig:comparison_DBN}
    \end{subfigure}
    \hfill
    \begin{subfigure}[b]{0.328\textwidth}
        \centering
        \includegraphics[width=\textwidth]{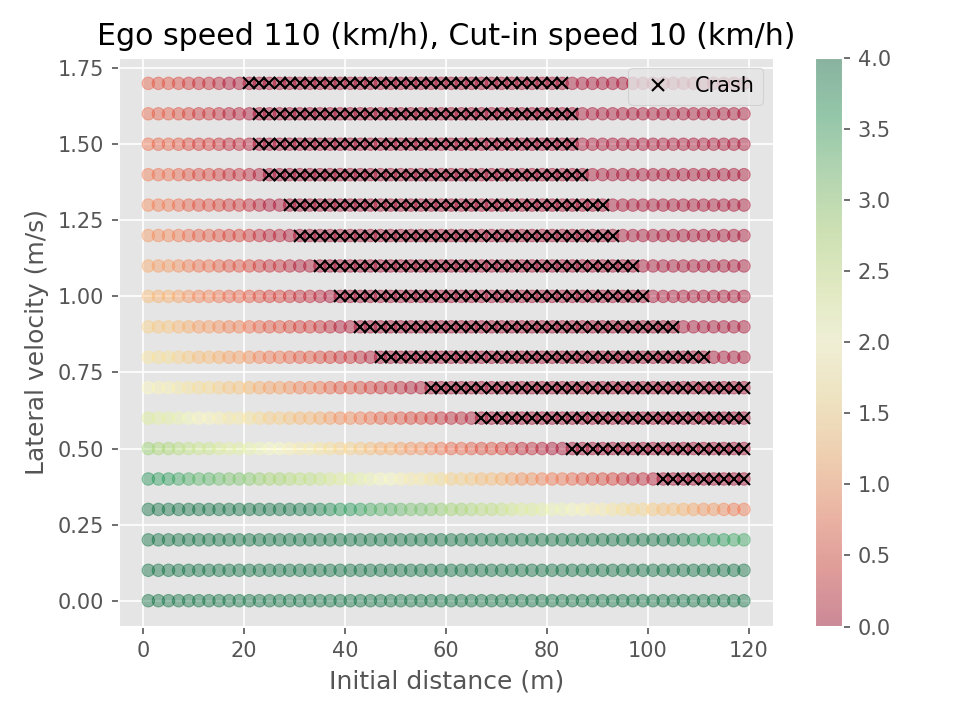}
        \caption{CC\_human\_driver}
        \label{fig:comparison_CC}
    \end{subfigure}
    \hfill
    \begin{subfigure}[b]{0.328\textwidth}
        \centering
        \includegraphics[width=\textwidth]{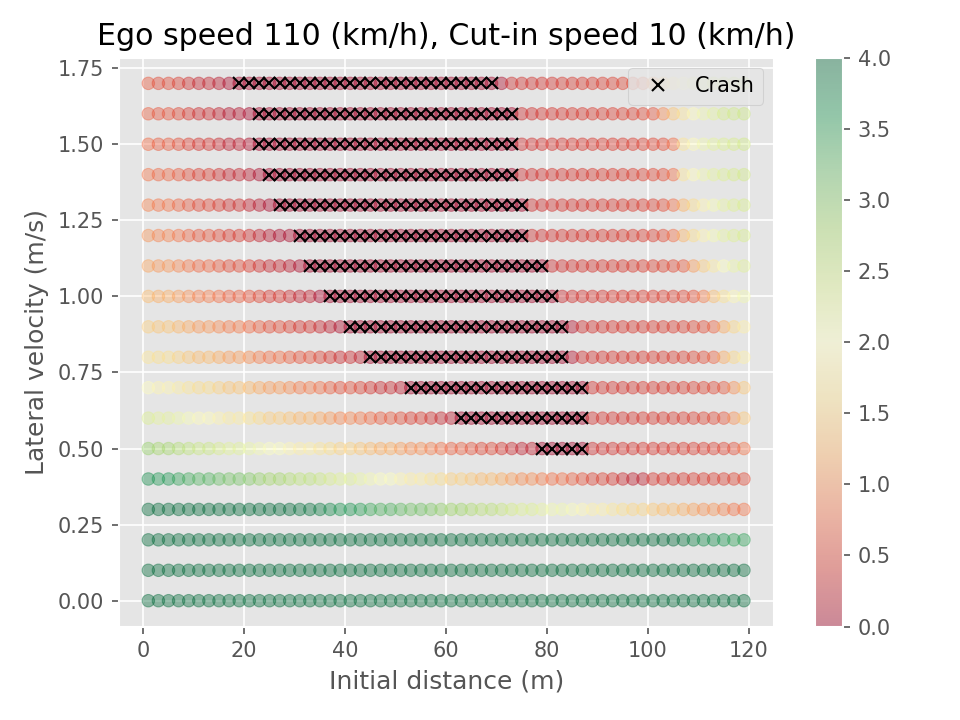}
        \caption{FSM}
        \label{fig:comparison_FSM}
    \end{subfigure}
    \\[1em] 
    \begin{subfigure}[b]{0.328\textwidth}
        \centering
        \includegraphics[width=\textwidth]{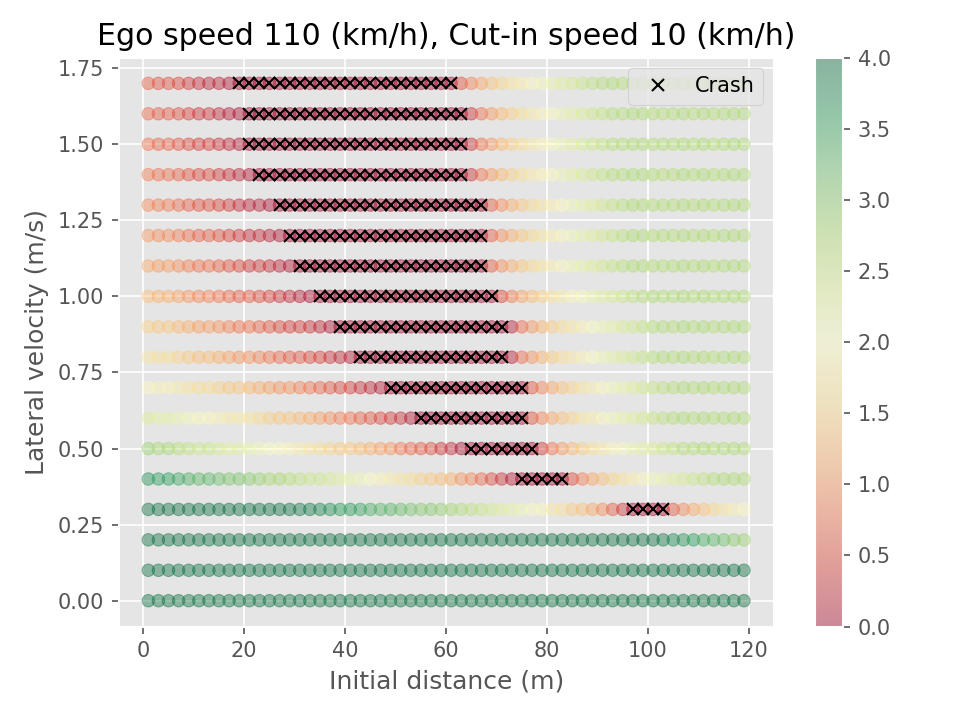}
        \caption{RSS}
        \label{fig:comparison_RSS}
    \end{subfigure}
    \hspace{0.1\textwidth} 
    \begin{subfigure}[b]{0.328\textwidth}
        \centering
        \includegraphics[width=\textwidth]{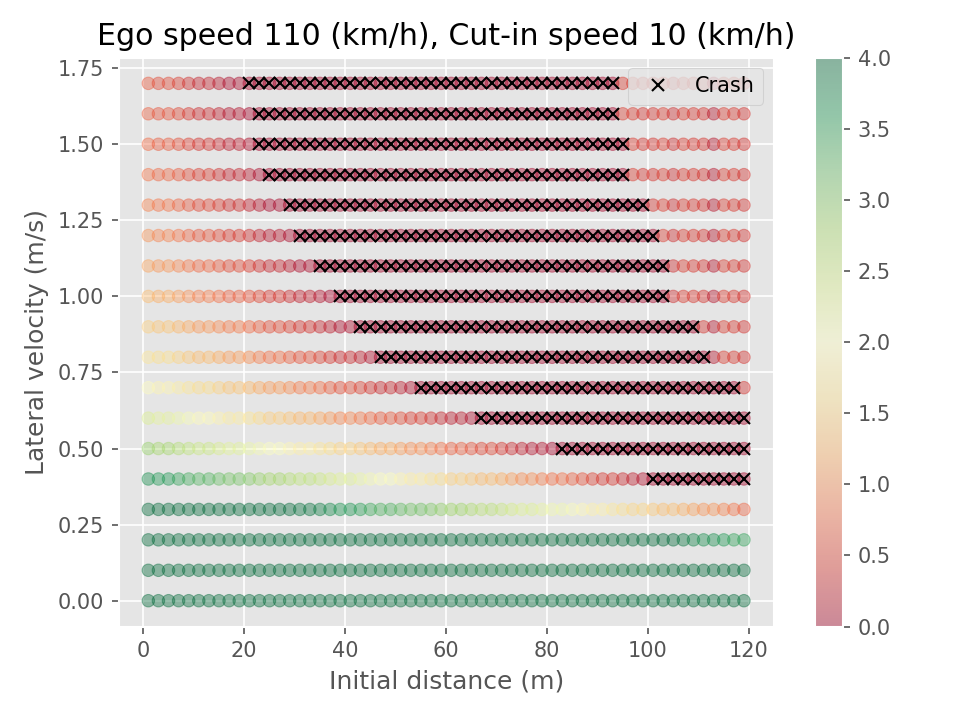}
        \caption{Reg157}
        \label{fig:comparison_Reg157}
    \end{subfigure}
    \caption{Performance comparison of all models at a High-Speed scenario.}
    \label{fig:performance_comparison}
\end{figure*}

As previously discussed, the simulator comprises five models: DBN, CC, FSM, RSS, and Reg157. This sub-section briefly compares and analyzes the DBN performance against all other existing models. Figures \ref{fig:comparison_DBN}, \ref{fig:comparison_CC}, \ref{fig:comparison_FSM}, \ref{fig:comparison_RSS}, and \ref{fig:comparison_Reg157} show all the model's performance for an extremely critical situation with an ego vehicle at 110 km/h and a cut-in vehicle approaching 10 km/h across all combinations of lateral speed and initial/front distance. Black cross marks indicate crashes, and non-crash scenarios with different colors, using a heatmap, indicate the TTC, which ranges from 0 to 4 (with green indicating a high TTC and red indicating a low TTC). From the Figures, it is clear that, especially in this scenario, DBN has the fewest crashes and better TTC compared to other models. Moreover, RSS and FSM models exhibit moderate crashes in comparison to DBN.

\begin{figure}[h]
    \centering
    \includegraphics[width=1\linewidth, height=0.22\textheight]{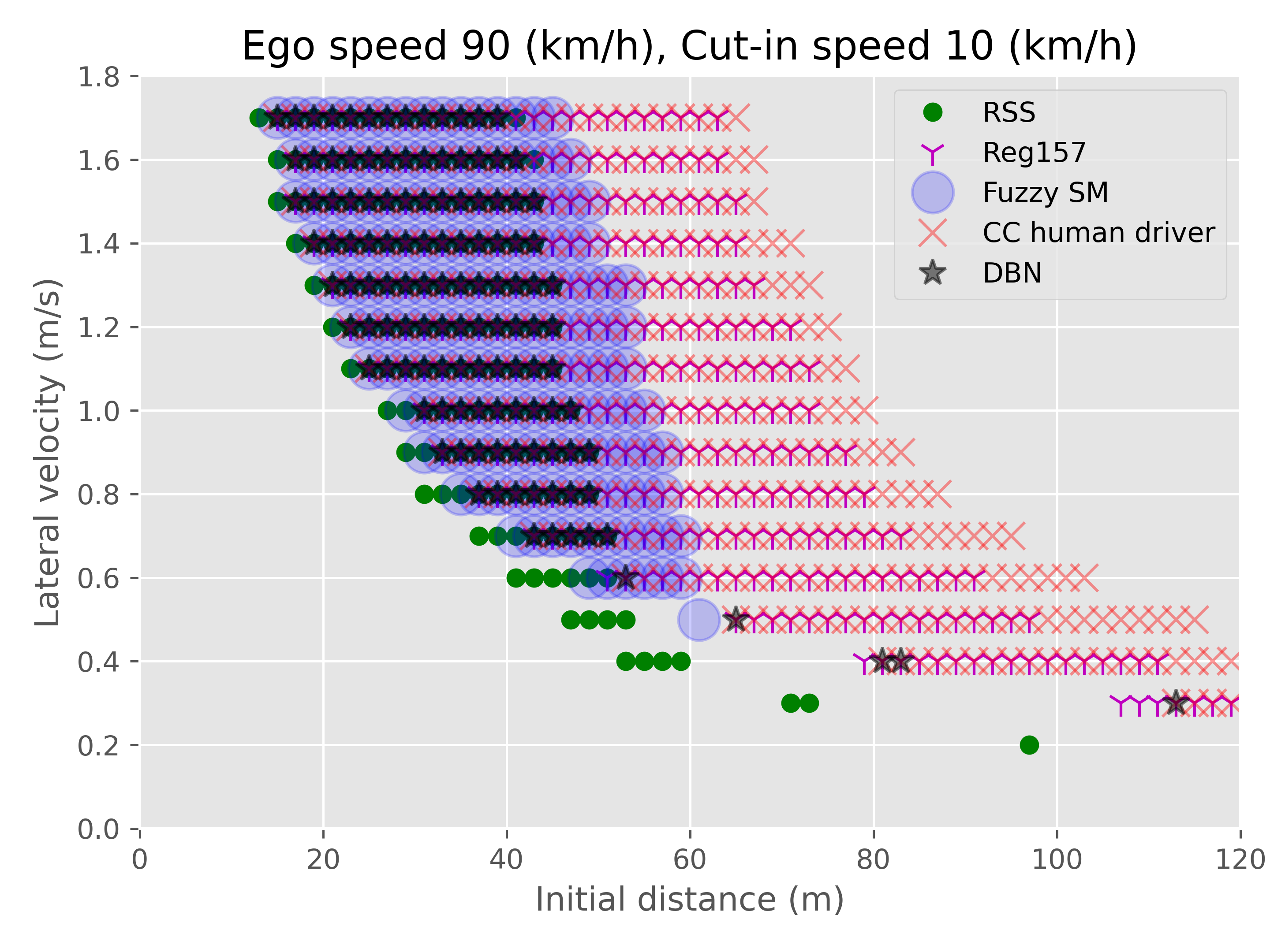} 
    \caption{Combined performance comparison of all models.}
    \label{fig:comparison_for_speeds_90_10} 
\end{figure}

To compare the effectiveness of all the models in a single plot, as in Figure \ref{fig:comparison_for_speeds_90_10}, we considered another critical scenario with ego and cut-in vehicle speeds set at 90 km/h and 10 km/h, respectively. Each model’s outcome is depicted through different marker types and colors. The plot demonstrates that DBN managed to avoid crashes in most cases, whereas FSM and RSS are comparatively close to DBN's performance. On the contrary, CC\_human\_driver and Reg157 are the least performing with a larger number of crashes.    

Furthermore, Table \ref{tab:crash_percentage} shows the overall crash percentage averaged across all the combinations of high-speed scenarios, ranging from 70 km/h to 130 km/h, and low speeds, from 10 km/h to 60 km/h. In the case of high-speed scenarios, DBN shows superior performance with the lowest crash percentage of 9.22\%. Baseline models like CC\_human\_driver and Reg157 show the highest crash percentages of 25.30\% and 20.43\%, respectively. The performance of DBN is comparable with FSM and RSS in low-speed use cases, with 6.81\% of the crashes. This demonstrates the effectiveness of DBN in reducing crashes at high-speed scenarios, where proactive detection and safety checks are crucial.

\begin{table}[h!]
\renewcommand{\arraystretch}{1.1}
\centering
\caption{Crash Percentage by Models for all Scenarios}
\label{tab:crash_percentage}
\begin{tabular}{lccc}
\hline
\textbf{Model} & \textbf{Low Speed (\%)} & \textbf{High Speed (\%)} \\ \hline
RSS & \textbf{5.52} & 10.62 \\
Reg157 & 14.14 & 20.43 \\
Fuzzy SM & 5.66 & 11.75 \\
CC\_human\_driver & 23.40 & 25.30 \\
DBN & 6.81 & \textbf{9.22} \\ \hline
\end{tabular}
\end{table}

\subsection{Statistical Insights}

The statistical analysis in Table \ref{tab:model_performance} focuses exclusively on extreme cases, specifically scenarios with high relative speeds where models are most prone to crashes: 90 km/h with 10 km/h and 130 km/h with 10 km/h for high speeds, and 30 km/h with 10 km/h and 60 km/h with 10 km/h for low speeds. These cases were selected to stress-test the models under the most challenging conditions, highlighting their performance in critical safety scenarios. Despite the wide range of possible configurations, these combinations effectively reveal the crash tendencies and limitations of the models. The analysis underscores DBN’s capability to reduce crashes significantly, achieving a crash percentage as low as 11.48\% at 90 km/h compared to CC\_human\_driver’s 33.89\%. Statistical evaluations like this are crucial to ensure that models perform reliably in scenarios with heightened risk, providing insights into safety margins and informed decision-making under stress.

\begin{table*}[ht]
\centering
\caption{Model Performance Statistics Across Some of the Critical Scenarios}
\renewcommand{\arraystretch}{1.05} 
\begin{tabular}{@{}cccccccc@{}}
\toprule
\textbf{Speed Range} & \textbf{Model} & \multicolumn{1}{c}{\textbf{Ego Speed}} & \multicolumn{1}{c}{\textbf{Cut-in Speed}} & \textbf{Number} & \textbf{Crash \%} & \textbf{Avg. TTC (s)} & \textbf{TTC Improvement} \\
& & \textbf{(km/h)} & \textbf{(km/h)} & \textbf{of Crashes} & & & \textbf{by DBN (s)} \\
\midrule
\multirow{10}{*}{\textbf{High}} & FSM & 90 & 10 & 170 & 15.74 & 2.09 & -0.01 \\
& CC\_human\_driver & 90 & 10 & 366 & 33.89 & 1.50 & \textbf{0.58} \\
& RSS & 90 & 10 & 155 & 14.35 & \textbf{2.34} & -0.26 \\
& Reg157 & 90 & 10 & 322 & 29.81 & 1.55 & 0.53 \\
& DBN & 90 & 10 & \textbf{124} & \textbf{11.48} & 2.08 & - \\
\cmidrule(lr){2-8}
& FSM & 130 & 10 & 412 & 38.15 & 1.99 & -0.09 \\
& CC\_human\_driver & 130 & 10 & 419 & 38.80 & 1.98 & \textbf{-0.07} \\
& RSS & 130 & 10 & 349 & 32.31 & 2.10 & -0.20 \\
& Reg157 & 130 & 10 & 464 & 42.96 & \textbf{2.12} & -0.22 \\
& DBN & 130 & 10 & \textbf{284} & \textbf{26.30} & 1.91 & - \\
\midrule
\multirow{10}{*}{\textbf{Low}} & FSM & 30 & 10 & \textbf{52} & 4.90 & 2.44 & 1.00 \\
& CC\_human\_driver & 30 & 10 & 231 & 21.75 & 0.77 & \textbf{2.68} \\
& RSS & 30 & 10 & 56 & 5.27 & 2.18 & 1.26 \\
& Reg157 & 30 & 10 & 133 & 12.52 & 0.86 & 2.58 \\
& DBN & 30 & 10 & 72 & \textbf{6.78} & \textbf{3.44} & - \\
\cmidrule(lr){2-8}
& FSM & 60 & 10 & 144 & 13.56 & \textbf{2.10} & -0.05 \\
& CC\_human\_driver & 60 & 10 & 438 & 41.24 & 1.75 & 0.31 \\
& RSS & 60 & 10 & 144 & 13.56 & 2.09 & -0.04 \\
& Reg157 & 60 & 10 & 304 & 28.63 & 1.55 & \textbf{0.51} \\
& DBN & 60 & 10 & \textbf{126} & \textbf{11.86} & 2.05 & - \\
\bottomrule
\end{tabular}
\label{tab:model_performance}
\end{table*}

\subsection{Qualitative Analysis}

\begin{figure}[h]
    \centering
    \includegraphics[width=1\linewidth, height=0.38\textheight]{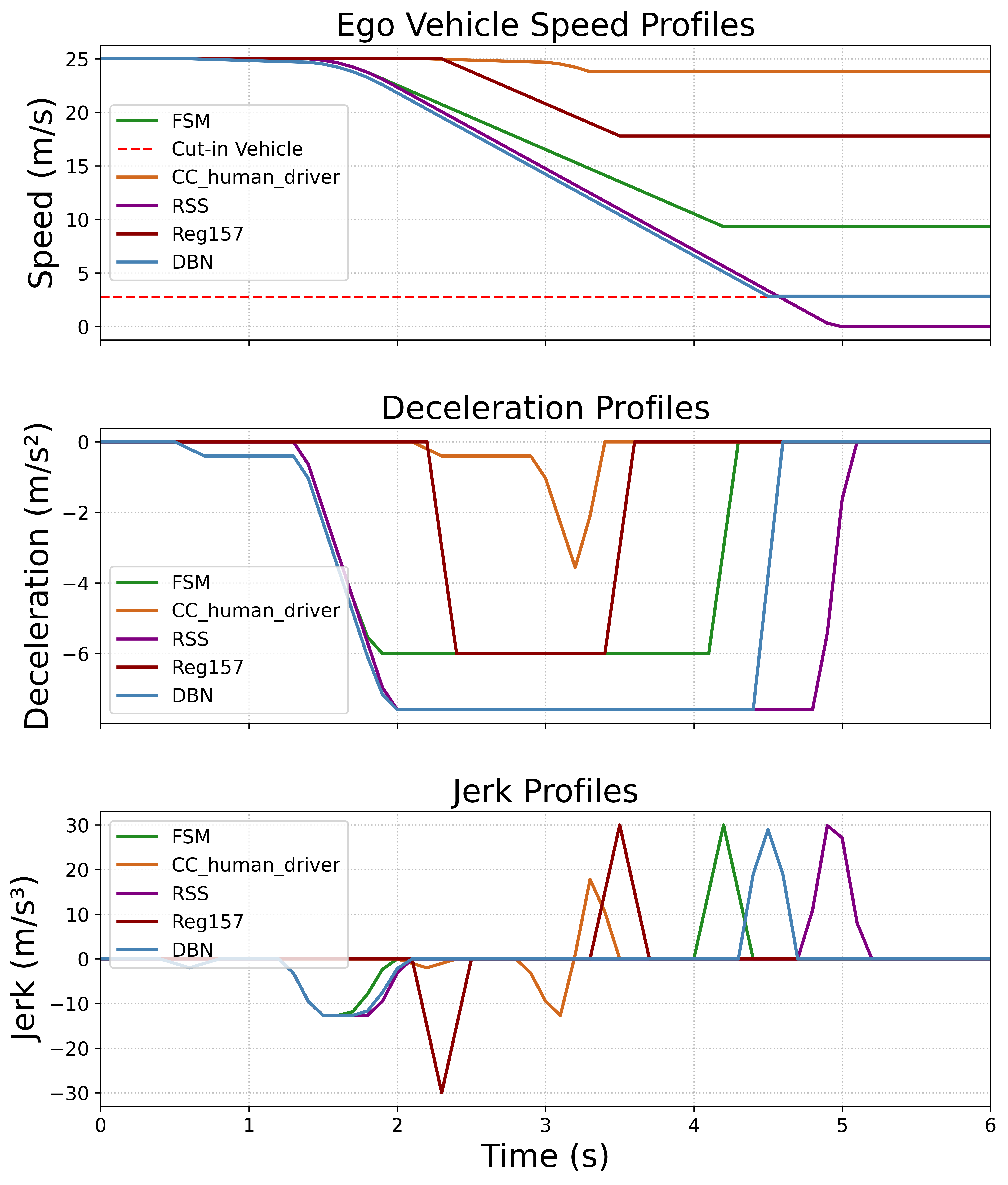} 
    \caption{Speed, deceleration, and jerk profiles of different models.}
    \label{fig:combined_profiles_new} 
\end{figure}

A deeper analysis of the capability of various models for effective early detection and smooth braking through seamless cut-in maneuvers is shown in Figure \ref{fig:combined_profiles_new}. A specific critical high-speed scenario is chosen, with an ego vehicle speed of 90 km/h, a cut-in vehicle speed of 10 km/h, a lateral speed of 1.7 m/s, and an initial longitudinal distance of 41 meters. This scenario was mainly chosen because the DBN model's efficient braking strategy allowed it to smoothly reduce speed and avoid the crash, while other models failed to do so. DBN is seen to facilitate the early detection of a lane change in the speed profile, which enables prompt and steady speed reduction, leading to successful crash avoidance. The deceleration profile demonstrates DBN's capability to apply controlled deceleration earlier, at 0.5 seconds, whereas the RSS and FSM models are applied at 1.3 seconds. DBN's performance is further highlighted in the jerk profile due to its better fluctuations when compared with huge oscillations of CC\_human\_driver and Reg157. More oscillations or jerks could result in higher instability and discomfort. This specific analysis shows the robustness and effectiveness of DBN over the other models in high-speed environments.

\section{Conclusion and Future Work}

This study proposes an enhanced DBN framework for estimating lane change and safe cut-in maneuvering in the JRC-FSM simulator. The proposed framework enhances the safety assessment through three major blocks of the DBN model, with a significant focus on integrating lateral evidence for safety checks into the CC\_human\_driver model. Integrating lane change detection into the decision-making pipeline proved advantageous in not only improving the performance over the baseline CC\_human\_driver model without causing additional crashes but also achieving superior performance when compared with all the other models (FSM, RSS, Reg157). 
In the simulator, the efficiency of this DBN was analyzed and tested under various critical speed ranges of ego and cut-in vehicles. Results show that DBN has outperformed other models with 9.2\% crashes in high-speed scenarios and maintained a competitive performance of 6.81\% crashes in low-speed scenarios. Future work will focus on enhancing the probabilistic reaction function within the DBN framework for adaptive speed control and proactive decision-making under dynamic traffic environments. This framework establishes a strong benchmark for analyzing the advantages of using probabilistic models for efficient risk management and accountability in the face of uncertainty. Moreover, the model's explainability makes it suitable for incorporation into AV safety regulations, ensuring safety validation and trusted decisions for advanced automated systems. 

\section{Acknowledgments}
This study has been partially funded by the European Commission Horizon Europe project “i4Driving” (Grant Agreement ID 101076165).

\bibliographystyle{IEEEtran}
\bibliography{references}

\end{document}